\pdfoutput=1

\documentclass[11pt]{article}

\usepackage{acl}

\usepackage{times}
\usepackage{latexsym}

\usepackage{graphicx}
\usepackage{subfigure}
\usepackage{amssymb,bm}
\usepackage{amsmath, amsthm}
\usepackage{multirow}
\usepackage{arydshln}

\usepackage{algorithm}
\usepackage{algorithmic}

\usepackage[T1]{fontenc}

\usepackage[utf8]{inputenc}

\usepackage{microtype}

\title{Robust Self-Augmentation for Named Entity Recognition with Meta Reweighting}

\author{
Linzhi Wu$^1$\thanks{\, Equal contributions.}\hspace{0.5em}, Pengjun Xie$^*$, Jie Zhou$^{2*}$, Meishan Zhang$^3$\thanks{\, Corresponding author.}\hspace{0.5em}, \\ \textbf{Chunping Ma, Guangwei Xu, Min Zhang$^3$} \\ 
$^1$School of New Media and Communication, Tianjin University \\ $^2$School of Electronic Information and Electrical Engineering, Shanghai Jiao Tong University 
\\ $^3$Institute of Computing and Intelligence, Harbin Institute of Technology (Shenzhen) \\ \small\texttt{\{tjuwlz2020, machunpingjj\}@163.com, sanny02@sjtu.edu.cn} \\
\small\texttt{\{xpjandy, mason.zms, ahxgwOnePiece\}@gmail.com, zhangmin2021@hit.edu.cn}
}

\begin{document}
\maketitle
\begin{abstract}
Self-augmentation has received increasing research interest recently to improve named entity recognition (NER) performance in low-resource scenarios. Token substitution and mixup are two feasible heterogeneous self-augmentation techniques for NER that can achieve effective performance with certain specialized efforts. Noticeably, self-augmentation may introduce potentially noisy augmented data. Prior research has mainly resorted to heuristic rule-based constraints to reduce the noise for specific self-augmentation methods individually. In this paper, we revisit these two typical self-augmentation methods for NER, and propose a unified meta-reweighting strategy for them to achieve a natural integration. Our method is easily extensible, imposing little effort on a specific self-augmentation method. Experiments on different Chinese and English NER benchmarks show that our token substitution and mixup method, as well as their integration, can achieve effective performance improvement. Based on the meta-reweighting mechanism, we can enhance the advantages of the self-augmentation techniques without much extra effort.
\end{abstract}


\section{Introduction}
Named entity recognition (NER), which aims to extract predefined named entities from a piece of unstructured text, is a fundamental task in the natural language processing (NLP) community,
and has been studied extensively for several decades  \cite{hammerton2003named,huang2015bidirectional,chiu2016named,ma2016end}.
Recently, supervised sequence labeling neural models have been exploited most popularly for NER,
leading to state-of-the-art (SOTA) performance \cite{zhang-yang-2018-chinese,li-etal-2020-flat,ma2020simplify}.



Although great progress has been made, developing an effective NER model usually requires a large-scale and high-quality labeled training corpus,
which is often difficult to be obtained in real-world scenarios due to the expensive and time-consuming annotations by human experts.
Moreover, it would be extremely serious because the target language, target domain, and the desired entity type could all be infinitely varied.
As a result, the low-resource setting with only a small amount of annotated corpus available is far more common in practice, even though it may result in significant performance degradation due to the overfitting problem.

Self-augmentation is a prospective solution to this problem,
which has received widespread attention \cite{zhang2018mixup,wei2019eda,dai-adel-2020-analysis,chen2020local,karimi2021aeda}.
The major motivation is to generate a pseudo training example set deduced from the original gold-labeled training data automatically.
For NER, a token-level task, the feasible self-augmentation techniques include token substitution \cite{dai-adel-2020-analysis,zeng2020counterfactual} and mixup \cite{zhang2020seqmix,chen2020local},
which are deformed at the ground-level inputs and the high-level hidden representations, respectively.

\begin{figure}[t]
\centering
\includegraphics[width=0.475\textwidth]{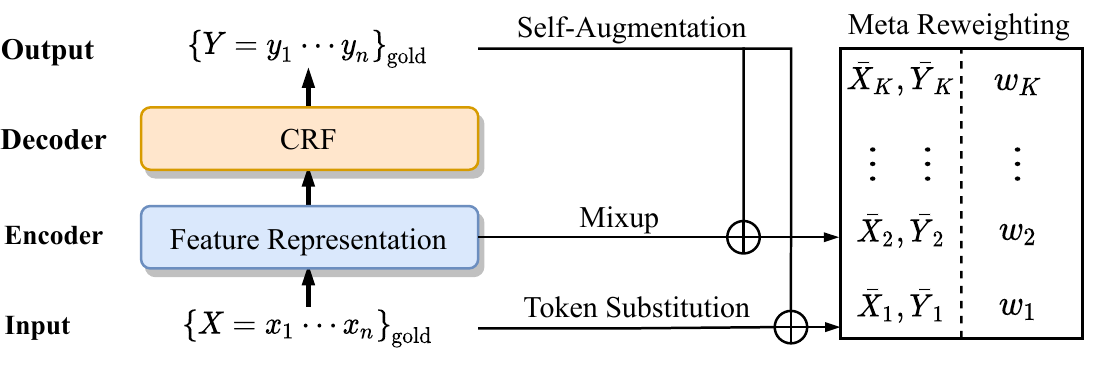}
\caption{The main idea of our work, where the two heterogeneous self-augmentation methods (i.e., token substitution and mixup) are integrated by a unified meta reweighting framework.}
\label{intro_arch}
\end{figure}



Nonetheless, there are still some limitations currently for the above token substitution and mixup methods.
For one thing, both of them require some specialized efforts to improve their effectiveness due to the potential noise introduced by the self-augmentation, which may restrict the valid semantic representation of the augmented data.
For instance, token substitution is typically limited to the named entities in the training corpus \cite{wu2019neural}, and the mixup tends to be imposed on the example pairs with small semantic distance gaps \cite{chen2020local}.
For another thing, though the two techniques seem to be orthogonal and probably complementary to each other, it remains a potential challenge to effectively and naturally integrate them.

In this work, we revisit the token substitution and mixup methods for NER,
and investigate the two heterogeneous techniques under a unified meta reweighting framework (as illustrated in Figure \ref{intro_arch}).
First, we try to relax the previous constraints to a broader scope for these methods,
allowing for more diverse and larger-scale pseudo training examples.
However, this would inevitably produce some low-quality augmented examples (i.e., noisy pseudo data) in terms of linguistic correctness, which may negatively affect the model performance.
To this end, we present a meta reweighting strategy for controlling the quality of the augmented examples and leading to noise-robust training.
Also, we can naturally integrate the two methods by using the example reweighting mechanism, without any specialization in a specific self-augmentation method.

Finally, we carry out experiments on several Chinese and English NER benchmark datasets to evaluate our proposed methods.
We mainly focus on the low-resource settings, which can be simulated by using only part of the standard training set when the scale is large. 
Experimental results show that both our token substitution and mixup method coupled with the meta-reweighting can effectively improve the performance of our baseline model, 
and the combination can bring consistent improvement. Positive gains become more significant as the scale of the training data decreases, indicating that our self-augmentation methods can  handle the low-resource NER well. In addition, our methods can still work even with a large amount of training data.
The code is available at \url{https://github.com/LindgeW/MetaAug4NER}.

\begin{figure*}[htp!]
	\centering
	\includegraphics[scale=0.95]{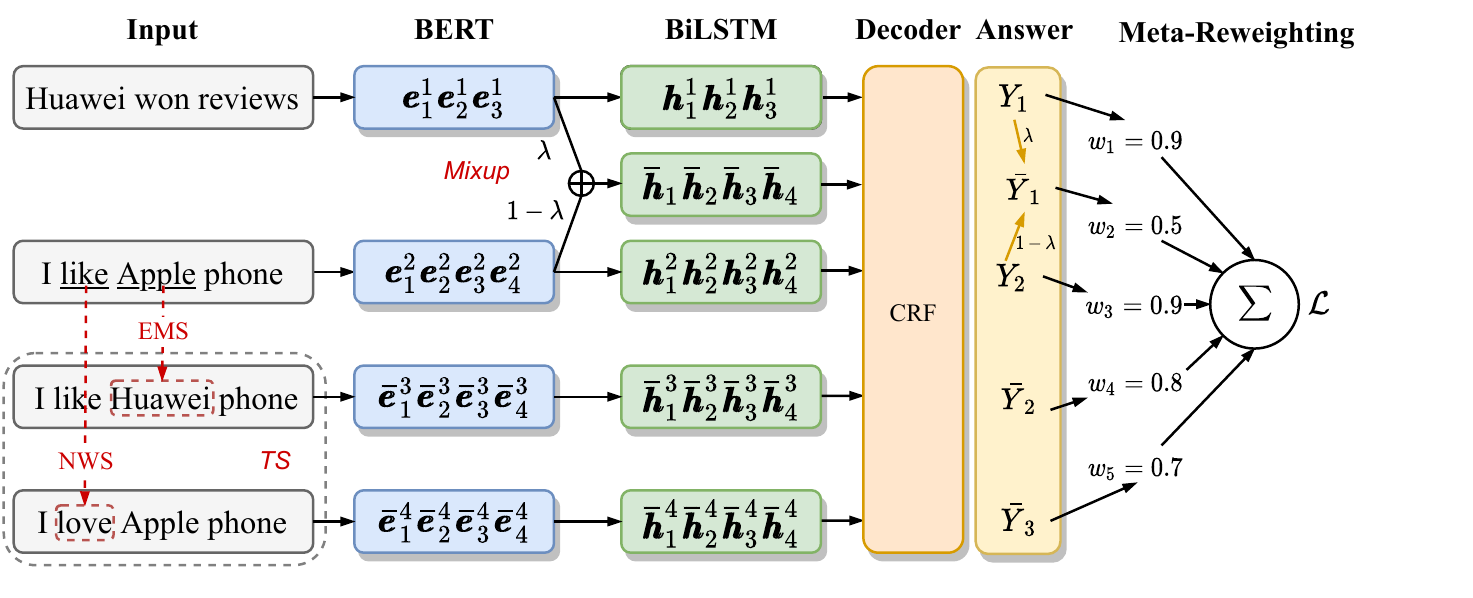}
    \caption{An overview of the self-augmentation framework for NER.}
	\label{fig:bert_lstm_crf}
\end{figure*}

\section{Our Approach}

In this section, we firstly describe our baseline model. Then, we present our self-augmentation methods to enhance the baseline model in the low-resource settings.
Finally, we elaborate on our meta reweighting strategy, which aims to alleviate the negative impact of the noisy augmented examples caused by the self-augmentation while also elegantly combining these augmentation methods.
\subsection{Baseline Model}
\label{baseline}
NER task is typically formulated as a sequence labeling problem, which transforms entities/non-entities into token-level boundary label sequence by using the BIO or BIOES schema \cite{huang2015bidirectional,lample2016neural}.
In this work, we adopt BERT-BiLSTM-CRF as our basic model architecture which consists of four components: (1) input representation,
(2) BiLSTM encoding, (3) CRF decoding, and (4) training objective.

\paragraph{Input Representation}
Given an input sequence $X = (x_1,\cdots ,x_n)$ of length $n$,
we first convert it into sequential hidden vectors using the pre-trained BERT \cite{devlin2019bert}:
\begin{equation}
    \label{bert_repr}
    \begin{split}
        \bm{e}_1,\cdots,\bm{e}_n=\text{BERT}(X),
    \end{split}
\end{equation}
where each token is mapped to a contextualized representation correspondingly.


\paragraph{Encoding}
We use a bidirectional LSTM layer to further extract the contextual representations,
where the process can be formalized as:
\begin{equation} \label{hidden}
    \begin{split}
        \bm{h}_1,\cdots ,\bm{h}_n &= \text{BiLSTM}(\bm{e}_1,\cdots ,\bm{e}_n), \\
    \end{split}
\end{equation}
where $\bm{h}_i$ is the hidden state output of the $i$-th token in the sequence ($i\in[1,n]$).

\paragraph{Decoding}
First, a linear transformation layer is used to calculate the initial label scores. Then, a label transition matrix $\bm{T}$ is used to model the label dependency.
Let $Y=(y_1, \cdots, y_n)$ be a label sequence,
the score $s(Y|X)$ can be computed by:
\begin{equation}\label{crf-score}
    \begin{split}
    & \bm{o}_i =\bm{W}\bm{h}_i+\bm{b}, \\
    & s(Y|X) = \sum_{i=1}^{n}(\bm{T}_{y_{i-1}, y_i}+\bm{o}_{i}[y_i]), \\
    \end{split}
\end{equation}
where $\bm{W}$, $\bm{b}$ and $\bm{T}$ are the model parameters.
Finally, we employ the Viterbi algorithm \cite{viterbi1967error} to find the best label sequence $Y$.

\paragraph{Training Objective}
We exploit the sentence-level cross-entropy objective for training.
Given a gold-labeled training example $(X, Y)$,
we have the conditional probability $p(Y|X)$ based on the scoring function defined in Equation \ref{crf-score},
and then apply a cross-entropy function to calculate the single example loss:
\begin{equation}\label{crf_loss}
    \begin{split}
        & p(Y|X) = \frac{\exp\big(s(Y|X)\big)}{\sum_{\widetilde{Y}}\exp\big(s(\widetilde{Y}|X)\big)},  \\
        & \mathcal{L}(X, Y) = -\log p(Y|X).
    \end{split}
\end{equation}
where $\widetilde{Y}$ denotes the candidate label sequences.


\subsection{Self-Augmentation}\label{des:aug}

Self-augmentation methods can reduce the demand for abundant manually-annotated examples, which can be implemented at the input level and representation level.
Token substitution and mixup are two popular methods for NER that correspond to these two distinct levels. 
Here, we try to extend these two self-augmentation methods.

\paragraph{Token Substitution}
\label{data_exp}
Token substitution aims to generate pseudo examples based on the original gold-labeled training data by replacing the tokens of input sentence with their synonym alternatives \cite{wu2019neural, dai-adel-2020-analysis}.
For NER, \citet{wu2019neural} adopted this method to obtain performance gains on Chinese datasets where the substituted objects are limited to named entities. \citet{dai-adel-2020-analysis} empirically demonstrated the superiority of synonym replacement among various augmentation schemes where the synonyms are retrieved from the off-the-shelf WordNet thesaurus.


Our token substitution is performed by building a synonym dictionary,
which covers the named entity synonyms as well as numerous normal word synonyms. Following \citet{wu2019neural}, we treat all entities of the same type from the training set as synonyms,
which are added to the entity dictionary.
We name it as entity mention substitution (EMS).
Meanwhile, we extend the substitution to non-entity tokens (i.e., the corresponding label is `O'), which is named as normal word substitution (NWS). Since unlabeled data in a specific domain is easily accessible, we adopt the word2vec-based algorithm \cite{mikolov2013distributed,pennington2014glove} to mine tokens with similar semantics on Wikidata via distributed word representation \cite{yamada2020wikipedia2vec}, and build a normal word synonym dictionary from the $k$-nearest token set based on cosine similarity distance. Note that this scheme does not require access to thesaurus for a specific domain in order to obtain synonyms.

Figure \ref{fig:bert_lstm_crf} presents an example of token substitution,
where EMS and NWS are both involved.
Specifically, for a given gold-labeled training example $(X, Y)$,
we replace the entity token of $X$ with a sampled entity from the entity dictionary which has the same entity type, and meanwhile replace the non-entity token of $X$ with a sampled synonym. 
Then, we can obtain a pseudo example $(\bar{X}, \bar{Y})$.
Especially, we balance the EMS and NWS strategies based on a ratio $\gamma$ by adjusting the percentage of EMS operations, aiming for a good trade-off between entity diversity and context diversity.
And, we refer to this method as TS in the rest of this paper for short.

\paragraph{Mixup for CRF}
Unlike token substitution performed at the ground input,
the mixup technique \cite{zhang2018mixup} generates virtual examples at the feature representation level in the NLP field \citep{guo2019augmenting}.
The main idea is to perform linear interpolations on both the input and ground-truth output between randomly sampled example pairs from the given training set.
\citet{chen2020local} presented the first work based on the token classification framework for the NER task, and the mixup strategy is constrained to the examples pairs where the input sentences are semantically similar by using specific heuristic rules. Different from their method, we extend the mixup technique to the CRF decoding.

Formally, give an example pair $(X_1, Y_1)$ and $(X_2, Y_2)$ randomly sampled from the gold-labeled training set, we firstly obtain their vector representations through Equation \ref{bert_repr},
resulting in $\bm{e}_{1,1}\cdots \bm{e}_{1,n_1}$, and $\bm{e}_{2,1}\cdots \bm{e}_{2,n_2}$, respectively.
Then we apply the linear interpolation to obtain a new virtual training example $(\bar{X}, \bar{Y})$.
Here we assume a regularization of pair-wise linear interpolation over the input representations and the output scores,
where the following attributes should be satisfied:
\begin{equation}
\label{mixup}
\begin{split}
&\bar{X}\text{:}\begin{cases}
&\text{BERT}(\bar{X}) = \bar{\bm{e}}_1\cdots \bar{\bm{e}}_n  \\
&\bar{\bm{e}}_i = \lambda \bm{e}_{1,i} + (1 - \lambda)\bm{e}_{2,i}, i \in [1,n]  \\
\end{cases} \\
&\bar{Y}\text{:} s(\bar{Y}|\bar{X}) = \lambda s(Y_1|\bar{X})  + (1 - \lambda) s(Y_2|\bar{X}) ,
\end{split}
\end{equation}
where $n = \max(n_1, n_2)$ \footnote{Special zero-vector pads are used to align two sequences with different lengths.} and $\lambda$ is sampled from a $\mathrm{Beta}(\alpha, \alpha)$
distribution $(\lambda \in [0, 1]~\text{and}~\alpha > 0)$. According to this formulation,
the loss function can be reformulated as:
\begin{equation}
    \begin{split}
        \mathcal{L}(\bar{X}, \bar{Y}) &= - \log \frac{\exp\big(s(\bar{Y}|\bar{X})\big)}{\sum_{\widetilde{Y}}\exp\big(s(\widetilde{Y}|\bar{X})\big)}
        \\ &= \lambda \mathcal{L}(\bar{X}, Y_1) + (1-\lambda) \mathcal{L}(\bar{X}, Y_2).
    \end{split}
\end{equation}
which aligns with the training objective of Equation \ref{crf_loss}.
In this way, our mixup method can fit well with the structural decoding.

\subsection{Meta Reweighting}
\label{diw_sec}

Although the self-augmentation techniques can efficiently generate numerous pseudo training examples, how to control the quality of augmented examples is a potential challenge that cannot be overlooked. In particular, unlike sentence-level classification tasks, entity recognition is highly sensitive to the semantics of the context. While positive augmented examples can help our model advance, some low-quality augmented examples that are inevitably introduced during self-augmentation may hurt the final model performance.


In this paper, we leverage a meta reweighting mechanism to dynamically and adaptively assign the example-wise weights to each mini-batch of training data, motivated by \citet{ren18l2rw}.
The key idea is that a small and clean meta-data set is applied to guide the training of model parameters, and the loss produced by the mini-batch of meta-data is exploited to reweight the augmented examples in each batch online. Intuitively, if the data distribution and gradient-descent direction of the augmented example are similar to those of the sample in the meta-data set, our model could better fit this \textit{positive} augmented sample and increase its weight, and vice versa. In other words, the clean and valid augmented examples are more likely to be fully trained.
\begin{algorithm}[t]
\caption{The training procedure of the meta reweighting strategy}
\label{alg:MR}
\begin{algorithmic}[1]
\REQUIRE Initial model parameters $\Theta^{(0)}$, clean training dataset $\mathcal{D}$, augmented training dataset $\mathcal{\hat{D}}$, batch size $m, n$, training steps $T$  \\
\ENSURE Updated model parameters $\Theta^{(T)}$
    \FOR{$t=1$ \textbf{to} $T$}
        \STATE Initialize the trainable parameter $\epsilon$.
        \STATE \{$x_c$, $y_c$\} $\gets$ SampleMiniBatch($\mathcal{D}$, $m$).
        \STATE \{$x_a$, $y_a$\} $\gets$ SampleMiniBatch($\mathcal{\hat{D}}$, $n$).
        \STATE $\mathcal{L}_a \gets \sum_{i=1}^{n}\epsilon_i \mathcal{L}(f(x_{a,i}; \Theta^{(t)}), y_{a,i})$.
        \STATE $\nabla\Theta^{(t)} \gets  \text{Grad}(\mathcal{L}_a, \Theta^{(t)})$.
        \STATE $\hat{\Theta}^{(t)} \gets \Theta^{(t)} - \beta\nabla\Theta^{(t)}$.
        \STATE $\mathcal{L}_c \gets \frac{1}{m}\sum_{i=1}^{m} \mathcal{L}(f(x_{c,i}; \hat{\Theta}^{(t)}), y_{c,i})$.
        \STATE $\nabla\epsilon \gets \text{Grad}(\mathcal{L}_c, \epsilon)$.
        \STATE $\hat{w} \gets \text{Sigmoid}(-\nabla\epsilon)$. 
        \STATE $w \gets \frac{\hat{w}}{\sum_{j}\hat{w}_j + \delta}$.
        \STATE $\hat{\mathcal{L}}_a \gets \sum_{i=1}^{n}w_i \mathcal{L}(f(x_{a,i}; \Theta^{(t)}), y_{a,i})$.
        \STATE $\nabla\Theta^{(t)} \gets \text{Grad}(\hat{\mathcal{L}}_a, \Theta^{(t)})$.
        \STATE $\Theta^{(t+1)} \gets \text{OptimizerStep}(\Theta^{(t)}, \nabla\Theta^{(t)})$.
    \ENDFOR
\end{algorithmic}
\end{algorithm}

More specifically, suppose that we have a set of $N$ augmented training examples $\hat{\mathcal{D}}=\{(X_i, Y_i)\}_{i=1}^N$, our final optimizing objective can be formulated as a weighted loss as follows:
\begin{equation}\label{opt_w}
    \begin{split}
        \Theta^*(w) = \mathop{\arg\min}\limits_{\Theta}\sum_{i=1}^N w_i\mathcal{L}(f(X_i; \Theta), Y_i),
    \end{split}
\end{equation}
where $w_i \ge 0$ is the learnable weight for the loss of $i$-th training example. $f(\cdot; \Theta)$ represents the forward process of our model (with parameter $\Theta$).
The optimal parameter $w$ is further determined by minimizing the following loss computed on the meta example set $\mathcal{D}=\{(X_i^m, Y_i^m)\}_{i=1}^M ~(M\ll N)$:
\begin{equation}\label{opt_w2}
    \begin{split}
       \!w^*\!=\!\mathop{\arg\min}\limits_{w}\!\frac{1}{M}\!\sum_{i=1}^M\!\mathcal{L}(f(X_i^m; \Theta^*(w)), Y_i^m),
    \end{split}
\end{equation}

Accordingly, we need to calculate the optimal $\Theta^*$ and $w^*$ in Equation \ref{opt_w} and \ref{opt_w2} based on two nested loops of optimization iteratively. For simplicity and efficiency, we take a single gradient-descent step for each training iteration to update them via an online-approximation manner.
At every training step $t$, we sample a mini-batch augmented examples $\{(X_i, Y_i)\}_{i=1}^{n}$ initialized with the learnable weights $\epsilon$. After a single optimization step, we have:
\begin{equation}
    \begin{split}
        \hat{\Theta}^{(t+1)}(\epsilon) = \Theta^{(t)} - \beta\nabla_{\small\Theta}\!\sum_{i=1}^{n}\epsilon_i\mathcal{L}(f(X_i;\Theta), Y_i),
    \end{split}
\end{equation}
where $\beta$ is the inner-loop step size. Based on the updated parameters, we then calculate the loss of the sampled mini-batch meta examples $\{(X_j^{meta}, Y_j^{meta})\}_{j=1}^{m}$:
\begin{equation}
    \begin{split}
        \!\mathcal{L}^{meta}(\hat{\Theta})\!=\!\frac{1}{m}\!\sum_{j=1}^{m}\mathcal{L}(f(X_j^{meta};\hat{\Theta}^{(t+1)}), Y_j^{meta}),
    \end{split}
\end{equation}
To generalize the parameters $\hat{\Theta}$ well to the meta-data set, we take the gradients of $\epsilon$ w.r.t the meta loss to produce example weights and normalize it along mini-batch:
\begin{equation}
    \begin{split}
        \hat{w}_i &= \sigma(-\nabla_{\epsilon_i}\mathcal{L}^{meta}(\hat{\Theta})\Big|_{\epsilon_i=0}), \\
        w_i &= \frac{\hat{w}_i}{\sum_{j}\hat{w}_j + \delta}.
    \end{split}
\end{equation}
where $\sigma(\cdot)$ is the sigmoid function and $\delta$ is a small value to avoid division by zero. Finally, we optimize the model parameters over augmented examples with the calculated weights. 

Algorithm \ref{alg:MR} illustrates the detailed training procedure of the meta reweighting strategy. It is noteworthy that the augmented training examples contain the original clean training examples, which serve as the unbiased meta-data.
Since the algorithm execution just requires a clear definition of the training objective for the input examples,
it is also well adaptable for the virtual augmented examples generated by our mixup method.

\begin{table*}[tb]
\centering
\setlength\tabcolsep{5pt}
\begin{tabular}{l|ccc|ccc|ccc}
\hline
\multirow{2}{*}{\textbf{Models}} & \multicolumn{3}{|c|}{\textbf{ON 4}} &
\multicolumn{3}{|c|}{\textbf{ON 5}} &
\multicolumn{3}{c}{\textbf{CoNLL03}}  \\ \cline{2-10}
& \textbf{5\%}      & \textbf{10\%}     & \textbf{30\%}     & \textbf{5\%}    & \textbf{10\%}   & \textbf{30\%} & \textbf{5\%}    & \textbf{10\%}    & \textbf{30\%}   \\ \hline
Baseline   &75.07   &76.14  &80.88 &81.22   &83.51  &86.27  & 85.12    & 87.11  & 89.24       \\  \hdashline
 + TS \textit{w/o} MR & 74.58  & 75.94  & 79.83  & 82.12    & 83.82     & 86.23  & 85.93    & 87.66    & 89.14  \\
 + TS \textit{w/} MR & 76.08   & 76.85  & 81.23 & 82.58    & 83.92    & 86.50  & 86.25   & 88.00    & 89.55   \\ \hdashline 
  + Mixup \textit{w/o} MR  & 75.21   & 76.03   & 80.00  & 82.63   & 83.77     & 86.04 & 86.18    & 87.75   & 89.48     \\
 + Mixup  \textit{w/} MR  &  76.15  & 76.75  & 80.97 & 82.83    &  84.12    & 86.60  &  86.33    & 88.03      &  89.75     \\ \hdashline 
 + Both \textit{w/o} MR  &  76.33 &  76.91  & 81.40  & 82.85  &  84.33  & 86.88  & 86.51   & 88.10  & 89.96  \\
 \bf + Both  (Final)  & \bf 76.82    & \bf 77.13    & \bf 81.66  & \bf 82.98    & \bf 84.52    & \bf 87.09  & \bf 86.76   & \bf  88.25  & \bf 90.12  \\\hline
\citet{dai-adel-2020-analysis} & 75.05 & 76.75 & 81.24 & 82.47 & 83.90 & 86.55 & 86.22 & 87.86 & 89.91 \\
\citet{chen2020local} & -- & -- & -- & -- & -- & -- & 84.85 & 87.85 & 89.87  \\ 
\citet{chen2020local} (\textit{Semi}) & -- & -- & -- & -- & -- & -- & 86.33 & 88.78 & 90.25   \\ 
\hline
\end{tabular}
\caption{Results on OntoNotes and CoNLL03 using 5\%, 10\%, and 30\% of the training data. \textit{Semi}: additional 10,000 unlabeled training examples are used.}
\label{tab:limit_onto_conll}
\end{table*}


\begin{table}[tb]
\centering
\setlength\tabcolsep{10pt}
\begin{tabular}{l|c|c}
\hline
\textbf{Models} &
\textbf{ON 4} & \textbf{Weibo} \\    \hline
Baseline &81.73  & 69.10  \\\hdashline
+ TS \textit{w/o} MR &81.38   &68.69  \\
+ TS \textit{w/} MR & 81.85   &69.61  \\ \hdashline
+ Mixup \textit{w/o} MR & 81.68   &69.96  \\
+ Mixup \textit{w/} MR &  82.15  & 70.53  \\
\hdashline 
+ Both \textit{w/} MR & 82.33 & 71.15 \\
\bf + Both (Final) &\bf 82.48   & \bf 71.42 \\
\hline
\citet{meng2019glyce}$^\dagger$ &81.63  & 67.60       \\
\citet{hu2020slk}  &80.20  & 64.00              \\
\citet{xueporous}  &80.60   &69.23 \\
\citet{li-etal-2020-flat} &81.82  & 68.55        \\
\citet{nie2020improving}$^\dagger$  &81.18  & 69.78        \\
\citet{nie2020named} &--   & 69.80        \\
\citet{li-etal-2020-unified}$^\dagger$ &82.11  &-- \\
\citet{ma2020simplify} & 82.81  & 70.50       \\
\citet{xuan2020fgn}$^\dagger$ &82.04  & 71.25   \\
\citet{liu-etal-2021-lexicon} & 82.08 & 70.75 \\
\hline
\end{tabular}
\caption{Performance comparisons using the full training data on  OntoNotes 4 (ON 4) and Weibo. Previous SOTA results are also offered for comparisons. $\dagger$ denotes external knowledge is used.}
\label{tab:full_onto}
\end{table}

\begin{table}[tb]
\centering
\setlength\tabcolsep{5pt}
\begin{tabular}{l|c|c}
\hline
\textbf{Models} & \textbf{CoNLL03} & \textbf{ON 5}  \\  \hline
Baseline   & 91.23 & 88.22 \\  \hdashline
 + TS \textit{w/o} MR & 90.98 & 87.55   \\
 + TS \textit{w/} MR & 91.64  & 88.84  \\ \hdashline
 + Mixup \textit{w/o} MR  & 91.04 & 87.46 \\ 
 + Mixup \textit{w/} MR  & 91.42  & 88.98 \\
 \hdashline 
 + Both \textit{w/} MR & 91.88 & 89.24 \\
 \bf + Both  (Final) & \bf 92.15 & \bf 89.43 \\\hline
\citet{chen2020local}  & 91.83  & -- \\
\citet{clark-etal-2018-semi}$^\ddagger$ & 92.60 & 88.80 \\
\citet{fisher-vlachos-2019-merge} & -- & 89.20 \\
\citet{li-etal-2020-unified}$^\dagger$ & 93.04 & 91.11  \\
\citet{yu-etal-2020-named}$^\dagger$ & 93.50 & 91.30  \\
\citet{xu2021better}$^\dagger$ & -- & 90.85  \\
\hline
\end{tabular}
\caption{Performance comparisons using the full training data on CoNLL03 and OntoNotes 5 (ON 5). Previous SOTA results are also offered for comparisons. $\ddagger$ means the multi-task learning with more unlabeled data. $\dagger$ denotes external knowledge is used.}
\label{tab:full_conll}
\end{table}


\section{Experiments}
\subsection{Settings}
\paragraph{Datasets} To validate our methods, we conduct experiments on Chinese benchmarks: OntoNotes 4.0 \cite{weischedel2011ontonotes} and Weibo NER \cite{peng2015named}, as well as English benchmarks: CoNLL 2003 \cite{sang2003introduction} and OntoNotes 5.0\footnote{\url{https://catalog.ldc.upenn.edu/LDC2013T19}} \cite{pradhan-etal-2013-towards}.
The Chinese datasets are split into training, development and test sections following \citet{zhang-yang-2018-chinese} while we take the same data split as \citet{benikovagermeval} and \citet{conll20120shared} on the English datasets.
We follow \citet{lample2016neural} to use the BIOES tagging scheme for all datasets. 
The detailed statistics can be found in Table \ref{data_summary}.

\begin{table}[H]
\setlength\tabcolsep{3.8pt}
\centering
\begin{tabular}{llllll}
\hline
\textbf{Dataset} & \textbf{Type}   & \textbf{Train} & \textbf{Dev} & \textbf{Test} \\ \hline
\multirow{3}{*}{OntoNotes 4} & \#sent & 15.7k &4.3k     &4.3k      \\
& \#char & 491.9k &200.5k  &208.1k  \\   
& \#entity & 12.6k & 6.6k & 7.3k \\ \hline
\multirow{3}{*}{Weibo}     & \#sent &1.4k   &0.27k     &0.27k      \\
&  \#char &73.8k &14.5k &14.8k      \\
& \#entity & 1.9k & 0.4k & 0.4k \\ \hline
\multirow{3}{*}{CoNLL03}  & \#sent & 15.0k & 3.5k & 3.7k \\
& \#token & 203.6k & 51.4k & 46.4k \\ 
& \#entity & 23.5k & 5.9k & 5.6k \\ \hline
\multirow{3}{*}{OntoNotes 5}  & \#sent & 59.9k & 8.5k & 8.3k \\
& \#token & 1088.5k & 147.7k & 152.7k \\
& \#entity & 81.8k & 11.1k & 11.3k \\\hline
\end{tabular}
\caption{Statistics of datasets. \#sent and \#entity stand for the number of sentences and entity words, respectively.}
\label{data_summary}
\end{table}

\paragraph{Implementation Details} 
We use one-layer BiLSTM and the hidden size is set to 768.
The dropout ratio is set to 0.5 for the input and output of BiLSTM.
Regarding BERT, we adopt BERT-base model\footnote{\url{https://github.com/huggingface/transformers}} (BERT-base-cased for the English NER) and fine-tune the inside parameters together with all other module parameters.
We use the AdamW\cite{Loshchilov2019DecoupledWD} optimizer to update the trainable parameters with $\beta_1$=0.9 and $\beta_2$=0.99. For the BERT parameters, the learning rate is set to $2e{-}5$. For other module parameters excluding BERT, a learning rate of $1e{-}3$ and weight decay of $1e{-}4$ are used. Gradient clipping is used to avoid gradient explosion by a maximum value of 5.0. 
All the models are trained on NIVIDIA Tesla V100 (32G) GPUs. The \texttt{higher} library\footnote{\url{https://github.com/facebookresearch/higher}} is utilized for the implementation of second-order optimization involved in Algorithm \ref{alg:MR}.

For the NWS, we use the word vectors trained on  Wikipedia data\footnote{\url{https://dumps.wikimedia.org/}} based on the GloVe model \cite{pennington2014glove} and build the synonym set for any given non-entity word based on the top-5 cosine similarity, where stop-words are excluded.
As mentioned in Section \ref{des:aug}, we defined two core hyper-parameters for our self-augmentation methods, one for TS (i.e., $\gamma$) and the other for mixup  (i.e., $\lambda$).
Specifically, we set $\gamma = 20\%$ and $\lambda$ by sampled from the $\mathrm{Beta}(\alpha, \alpha)$ distribution with  $\alpha=7$,
where the details will be shown in the analysis section. 
Meanwhile, we conduct the augmentation up to 5 times corresponding to the original training data.
\paragraph{Evaluation} We conduct each experiment by 5 times and report the average F1 score. The best-performing model on the development set is then used to evaluate on the test set.

\subsection{Main Results}
The main results are presented in Table \ref{tab:limit_onto_conll}, \ref{tab:full_onto} and \ref{tab:full_conll},
verifying the effectiveness of our method under the low-resource setting and the standard full-scale setting, respectively. Since Weibo is a small-scale dataset, we do not consider its partial training set for the low-resource setting.

\paragraph{Low-Resource Setting}
We randomly sample  5\%, 10\%, and 30\% of the original training data from OntoNotes and CoNLL03 for simulation studies.
Table \ref{tab:limit_onto_conll} shows the results where F1 scores of the baseline, +TS, +Mixup, and +Both are reported.
We can observe that: (1) the baseline performance will drop significantly as the size of training data get reduced gradually, which demonstrates the performance of the supervised NER model relies heavily on the scale of the labeled training data. (2) although the number of training examples has increased, vanilla self-augmentation (without meta reweighting) might degrade the model performance due to potentially unreliable pseudo-labeled examples. 
The meta reweighting strategy helps to adaptively weight the augmented examples during training, which combats the negative impact and leads to a stable and positive performance boost.



In addition, as the scale of the training data decreases, the effectiveness of the augmentation methods can be more significant, indicating that our self-augmentation methods are highly beneficial for the low-resource settings, and the two-stage combination of the two heterogeneous methods can yield better performance consistently.

\paragraph{Full-Scale Setting}
Table \ref{tab:full_onto} and \ref{tab:full_conll} show the results using full-scale training data. The results demonstrate that our baseline model is already strong. The model after vanilla augmentation could perform slightly worse since each training example is treated equally even if it is noisy. This also implies our meta reweighting makes great sense.
Furthermore, our final model (+Both) can further achieve performance gains by integrating these self-augmentation methods with the meta reweighting mechanism. The overall trend is similar to the low-resource setting, but the gains are relatively smaller when the training data is sufficient. That may be attributed that the size of training data is large enough to narrow the performance gap between the baseline and augmented models. It also suggests that our method does not hurt the model performance even when using enough training data.



\paragraph{Comparison with Previous Work}
We also compare our method with previous representative SOTA work, where all referred systems exploit the pre-trained BERT model.
As shown, compared to \citet{dai-adel-2020-analysis} and \citet{chen2020local}, our method performs better when using limited training data. For \citet{chen2020local}, the pure mixup performs slightly better due to the well-designed example sampling strategy, but our overall framework outperforms theirs. Moreover, our method can match the performance of the semi-supervised setting that uses additional 10K unlabeled training data.
Besides, our final model, without utilizing much external knowledge, can achieve very competitive results on the full training set in comparison to most previous systems. 


\subsection{Analysis}
In this subsection, we further conduct detailed experimental analyses
on the CoNLL03 dataset for a better understanding of our method.
Our main concern is on the low-resource setting,
therefore the models based on 5\%, 10\% and 30\% of the original training data are our main focus.

\begin{figure}[!htbp]
\centering
\includegraphics[width=0.47\textwidth]{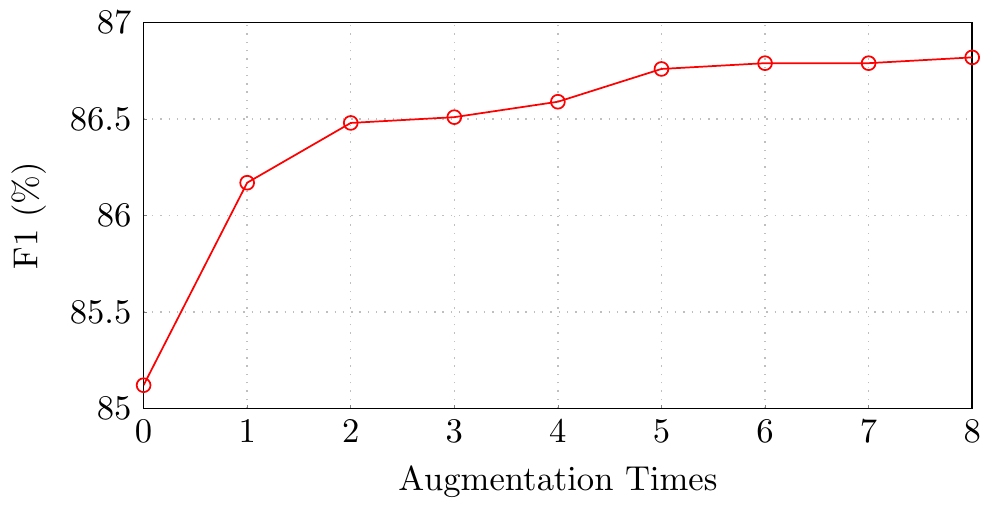}
\caption{Influence of augmentation times for 5\% training data of CoNLL03. Times=0 means original training data without any augmentation.}
\label{fig:aug_times}
\end{figure}

\paragraph{Augmentation Times}
The size of augmented examples is an essential factor in final model performance. Typically, we examine the 5\% CoNLL03 training data. As illustrated in Figure \ref{fig:aug_times}, the larger pseudo examples can obtain better performance in a certain range. However, as the times of augmentation increases, the uptrend of performance slows down. The improvement tends to be stable when the pseudo samples are increased 
to about 5 times the original training data. Excessively increasing the augmentation times does not necessarily bring consistent performance improvement. And we select an appropriate value for training data of different sizes from a range [1, 8].

\begin{figure}[t]
\centering
\includegraphics[scale=0.72]{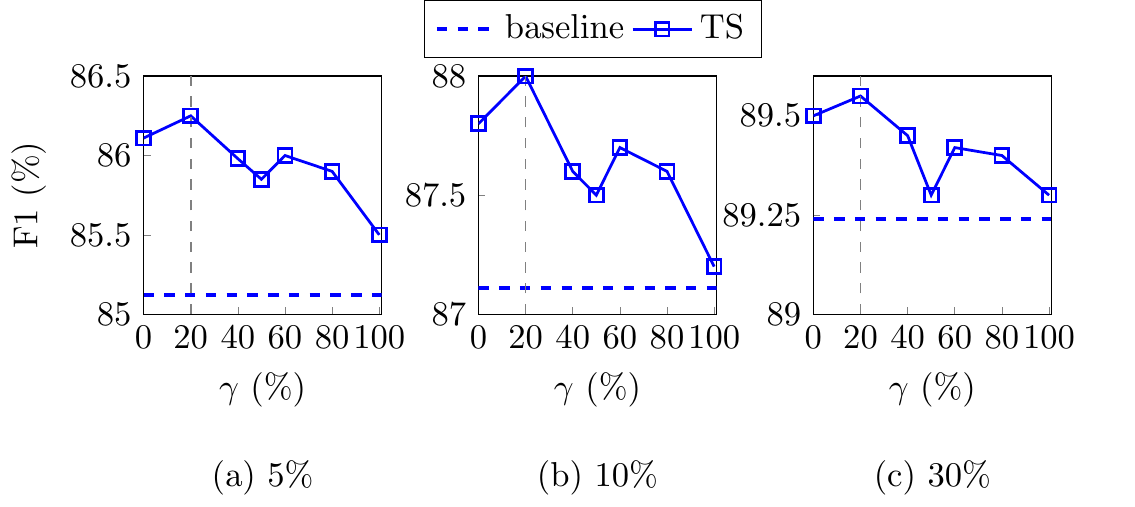}
\caption{Performance against different EMS rate $\gamma$.}
\label{ems_rate}
\end{figure}

\paragraph{Influence of $\gamma$ for Token Substitution}
Regarding our TS strategy, we take both NWS and EMS into account simultaneously. The two parts are blended by a percentage parameter $\gamma$, namely $\gamma$ for EMS and $1-\gamma$ for NWS.
Here we examine the influence of $\gamma$ in the sole self-augmentation model by TS.
Figure \ref{ems_rate} shows the results, where $\gamma = 0$ and $\gamma = 100\%$ denote the model with only NWS and EMS, respectively.
As shown, our model can achieve the overall better performance when $\gamma = 20\%$,
indicating that both of them are helpful for the TS strategy,
and NWS can be slightly better. One possible reason is that the entity words in original training examples are relatively sparse (i.e., the `O' label is dominant), allowing the NWS to produce more diverse pseudo examples.

\begin{figure}[t]
\centering
\includegraphics[scale=0.74]{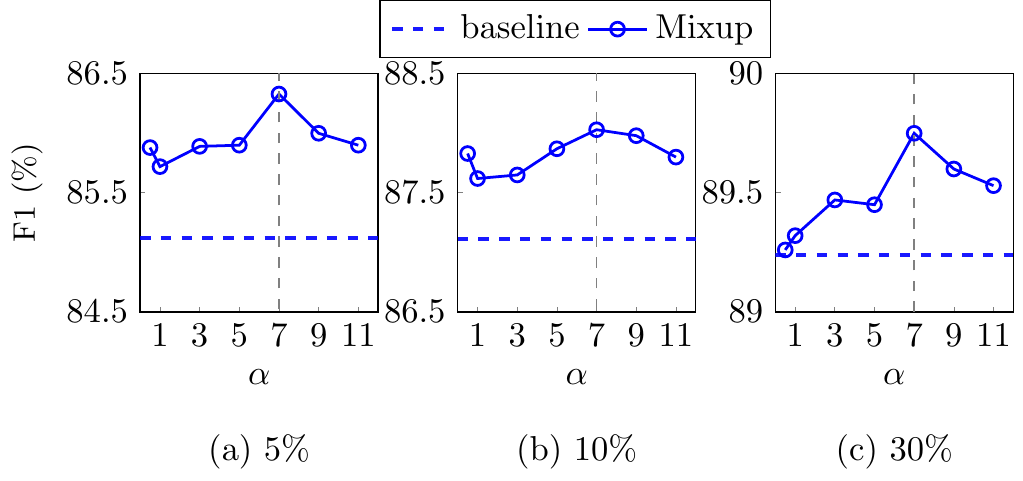}
\caption{Performance against different mixing parameter by $\mathrm{Beta}(\alpha, \alpha)$ distribution.}
\label{alpha_select}
\end{figure}

\paragraph{Mixup Parameter}
We further inspect the model with the mixup strategy alone so as to understand the important factors of the mixup model.
First, we analyze the influence of the mixing parameter $\alpha$.
As depicted in Figure \ref{alpha_select},
we can see that  $\alpha$ indeed affects the effectiveness of the mixup method greatly. Considering the feature of Beta distribution, the sampled $\lambda$ will be more concentrated around 0.5 as the $\alpha$ value becomes large, resulting in a relatively balanced weight between the mixed example pairs. The model performance remains stable when $\alpha$ is around 7.
Second, we study where to conduct the mixup operation since there are two main options in our framework, i.e., the hidden representations of either the BERT or BiLSTM for linear interpolation.
Table \ref{mix_layer} reports the comparison results, demonstrating the former is a better choice.

\begin{table}[!htbp]
\centering
\begin{tabular}{l|l|l|l}
\hline
Method       & \textbf{5\%} & \textbf{10\%} & \textbf{30\%} \\ \hline
Baseline      & 85.12    & 87.11  & 89.24  \\
Mixing on BiLSTM  & 86.15 & 87.66 &  89.41    \\
Mixing on BERT & \textbf{86.33}    & \textbf{88.03}      &  \textbf{89.75}      \\ \hline
\end{tabular}
\caption{Performance comparison of the mixup strategy on BERT or BiLSTM layer.}
\label{mix_layer}
\end{table}

\paragraph{Case Study}
To further understand the effectiveness of the meta-reweighting mechanism,
we present several high-quality and low-quality examples in Table \ref{tab:cases}.
As shown, the difference between the positive and negative examples for TS could be reflected in the syntactic and semantic validity of the augmented examples. Similarly, for the mixup, it seems that the valid example pairs are more likely to generate positive augmented examples.
\begin{table}[t]
\centering
\begin{tabular}{l|p{4.2cm}}
\hline
\textbf{Augmentation} & \textbf{Examples} \\ \hline
Original &  [Diana]$_\text{PER}$ met [Will Carling]$_\text{PER}$ at an exclusive gymnasium in [London]$_\text{LOC}$.       \\ \hdashline
Positive TS  & [Freddy Pinas]$_\text{PER}$ invited [John Marzano]$_\text{PER}$ at an available gym in [UK]$_\text{LOC}$.     \\ \hdashline
Negative TS & [Tim Henman]$_\text{PER}$ visit [Simpson]$_\text{PER}$ at an available room in [NICE]$_\text{LOC}$. \\\hline
\multirow{2}{*}{Positive Mixup} & French 1997 budget due around September 10 - Juppe.        \\ \cline{2-2}
                          & Jewish 1999 deficit due about October 20 M.Atherton.        \\ \hline
\multirow{2}{*}{Negative Mixup}         & Olympic champion Agassi meets Karim Alami of Morocco in the first round.     \\ \cline{2-2}
                          & Olympic champion Nathalie Lancien of France also missed the winning attack.       \\ \hline
\end{tabular}
\caption{Case study on positive and negative augmentation with respect to the TS and mixup.}
\label{tab:cases}
\end{table}

\section{Related Work}
In recent years, research on NER has concentrated on either enriching input text representations \cite{zhang-yang-2018-chinese,nie2020named,ma2020simplify}
or refining model architectures with various external knowledge \cite{zhang-yang-2018-chinese,ye-ling-2018-hybrid,li-etal-2020-flat,xuan2020fgn,li-etal-2020-unified,yu-etal-2020-named,Shen2021LocateAL}. Particularly, NER model, with the aid of large pre-trained language models \cite{peters2018deep,devlin2019bert,roberta2019}, has achieved impressive performance gains. However, these models mostly depend on rich manual annotations, making it hard to cope with the low-resource challenges in real-world applications.
Instead of pursuing a sophisticated model architecture,
in this work, we exploit the BiLSTM-CRF model coupled with the pre-trained BERT as our basic model structure. 

Self-augmentation methods have been widely investigated in various NLP tasks \cite{zhang2018mixup,wei2019eda,dai-adel-2020-analysis,zeng2020counterfactual,ding-etal-2020-daga}.
The mainstream methods can be broadly categorized into three types: (1) token substitution \cite{kobayashi2018contextual,wei2019eda,dai-adel-2020-analysis,zeng2020counterfactual},
which performs local substitution for a given sentence,
(2) paraphrasing \cite{kumar2019submodular,xie2020unsupervised,zhang-etal-2020-parallel}, which involves sentence-level rewriting without significantly changing the semantics,
and (3) mixup \cite{zhang2018mixup, chen2020local,sun2020mixup}, which carries out the feature-level augmentation. As a data-agnostic augmentation technique, mixup can help improve the generalization and robustness of our neural model acting as an useful regularizer \cite{verma2019manifold}.
For NER, token substitution and mixup are very suitable and
have been exploited successfully with specialized efforts \cite{dai-adel-2020-analysis,chen2020local,zeng2020counterfactual},
while the paraphrasing strategy may result in structure incompleteness and token-label inconsistency, thus there has not been widely concerned yet.
In this work, we mainly investigate the token substitution and mixup techniques for NER, as well as their integration. Despite the success of various self-augmentation methods, quality control may be an issue easily overlooked by most methods.

Many previous studies have explored the example weighting mechanism in domain adaption \cite{jiang-zhai-2007-instance,wang-etal-2017-instance,osumi2019domain}. \citet{xia2018instance} and \citet{wang2019better} looked into the example weighting methods for cross-domain tasks. \citet{ren18l2rw} adapted the MAML algorithm \cite{finn2017model} and 
proposed a meta-learning algorithm to automatically weight training examples of the noisy label using a small unbiased validation set. Inspired by their work, we extend the meta example reweighting mechanism to the NER task, which is exploited to adaptively reweight mini-batch augmented examples during training. The main purpose is to mitigate the potential noise effects brought by the self-augmentation techniques, advancing a noise-robust model, especially in low-resource scenarios.

\section{Conclusion}
In this paper, we re-examine two heterogeneous self-augmentation methods (i.e., TS and mixup) for NER, extending them into more unrestricted augmentations without heuristic constraints. We further exploit a meta reweighting strategy to alleviate the potential negative impact of noisy augmented examples introduced by the aforementioned relaxation. Experiments conducted on several benchmarks show that our self-augmentation methods along with the meta reweighting mechanism are very effective in low-resource settings, and still work when enough training data is used.
The combination of the two methods can lead to consistent performance improvement across all datasets. Since our framework is general and does not rely on a specific model backbone, we will further investigate other feasible model structures. 

\section*{Acknowledgements}
We thank the valuable comments of all anonymous reviewers.
This work is supported by grants from the National Natural Science Foundation of China (No. 62176180).

\bibliography{anthology,new_custom}
\bibliographystyle{acl_natbib}

\end{document}